\documentclass[letterpaper, 10 pt, conference]{ieeeconf}
\IEEEoverridecommandlockouts
\usepackage{cite}
\usepackage{amsmath,amssymb,amsfonts}
\usepackage{algorithmic}
\usepackage{graphicx}
\usepackage{textcomp}
\usepackage{xcolor}
\usepackage{adjustbox}
\usepackage{hyperref}
\usepackage{booktabs}
\usepackage{subcaption}
\usepackage{multirow}
\usepackage{svg}
\usepackage[ruled]{algorithm2e}

\newcommand{\model}{CADENCE}
\newcommand{\smodel}{CADENCE }

\newcommand{\oldmodel}{GrAMMI }

\newcommand{\camera}[1]{\textcolor{black}{#1}}
\newcommand{\edit}[1]{\textcolor{black}{#1}}

\def\BibTeX{{\rm B\kern-.05em{\sc i\kern-.025em b}\kern-.08em
    T\kern-.1667em\lower.7ex\hbox{E}\kern-.125emX}}
\begin{document}



 

\title{\LARGE \bf
Diffusion Models for Multi-target Adversarial Tracking\
}

\author{Sean Ye$^{1}$, Manisha Natarajan$^{1}$, Zixuan Wu$^{1}$, and Matthew C. Gombolay$^{1}$
\thanks{*This work was supported in part by the Office of Naval Research (ONR) under grant numbers N00014-19-1-2076, N00014-22-1-2834, and N00173-21-1-G009, the National Science Foundation under grant CNS-2219755, and MIT Lincoln Laboratory grant number 7000437192.}
 \thanks{$^{1}$All authors are associated with the Institute of Robotics and Intelligent Machines (IRIM), Georgia Institute of Technology, Atlanta, GA 30308, USA.}%
}


\maketitle

\begin{abstract}

Target tracking plays a crucial role in real-world scenarios, particularly in drug-trafficking interdiction, where the knowledge of an adversarial target's location is often limited. Improving autonomous tracking systems will enable unmanned aerial, surface, and underwater vehicles to better assist in interdicting smugglers that use manned surface, semi-submersible, and aerial vessels. As unmanned drones proliferate, accurate autonomous target estimation is even more crucial for security and safety. This paper presents \underline{C}onstrained \underline{A}gent-based \underline{D}iffusion for \underline{EN}han\underline{CE}d Multi-Agent Tracking (\model), an approach aimed at generating comprehensive predictions of adversary locations by leveraging past sparse state information. To assess the effectiveness of this approach, we evaluate predictions on single-target and multi-target pursuit environments, employing Monte-Carlo sampling of the diffusion model to estimate the probability associated with each generated trajectory. We propose a novel cross-attention based diffusion model that utilizes constraint-based sampling to generate multimodal track hypotheses. Our single-target model surpasses the performance of all baseline methods on Average Displacement Error (ADE) for predictions across all time horizons.

\end{abstract}


\section{Introduction}


Unmanned aerial vehicles (UAVs) are extensively used in military and civilian applications, such as surveillance, search and rescue, anti-smuggling operations, wildlife tracking, and urban traffic monitoring \cite{uav_1}. These missions often involve tracking dynamic targets in large-scale environments, where predicting a target's current and future states is essential and internal states are not fully observable. However, tracking targets in complex environments, especially adversarial ones, presents significant challenges, including sparse observations and multiple possible future states for adversaries. As UAVs continue to advance and showcase their capabilities in diverse fields, refining methods for tracking dynamic targets is increasingly important.

While most works focus on single-target tracking in partially observable environments \cite{musicki07, xiao2015, prev_work}, the challenges becomes significantly more complex in multi-target tracking \cite{bose07}. One significant challenge is the need to maintain distinct \camera{probability distributions} for each target while correctly associating detections or observations with the different targets. Furthermore, maintaining multiple \camera{probability distributions} of various tracks in multi-target settings involves handling track fragmentation (track splitting and merging) when targets interact with one another.

Earlier methods for target tracking encompass model-based approaches like Particle Filters \cite{djuric2008target, rao2013visual, jia2016target} and Kalman Filters \cite{chen2000mixture, leven2009unscented}. However, these methods fail in sparse environments, where the vast majority of the time we do not receive any information on the target location. Model-free data-driven approaches \cite{ PANG2017406, LIU2020289}, can often outperform model-based approaches by estimating the behavior of the target with prior behavioral data rather than relying on expert-defined models. One model-free approach has shown promising results on this challenging task \cite{prev_work} by maximizing mutual information to regulate the components of a Gaussian Mixture Model. However, this model was limited to predicting a single time horizon and tracking a single target. Additionally the prior model is limited to a parametric formulation for the multimodal \camera{probability distributions} by using a mixture of Gaussians. 
\color{black}

In this work, we address single and multi-target tracking in large-scale pursuit environments using diffusion probabilistic models. Inspired by the recent success of diffusion models for trajectory generation in robotics \cite{chi2023diffusionpolicy, janner2022planning, liu2022structdiffusion}, we propose a novel approach for target track reconstruction under partial observability.
We design a novel approach named \textbf{C}onstrained \textbf{A}gent-based \textbf{D}iffusion for \textbf{EN}han\textbf{CE}d Multi-Target Tracking (\model) that employs cross-attention to enable information exchange across different agents. A key benefit of diffusion models is their non-parametric formulation for generating multimodal hypotheses as compared to prior work. Additionally, we take inspiration from the computer vision community and adapt the classifier-guided sampling formulation to steer the trajectory generation process to adhere to motion model and environmental constraints. 

\noindent \textbf{Contributions:} Our key contributions are:

\begin{itemize}

\item First, we propose \smodel to track multiple adversaries, utilizing a cross-attention based diffusion architecture that implicitly conducts target track assignment between the agents. 

\item Second, we propose a constraint-guided sampling process for our diffusion models to ensure that state transition functions and obstacle constraints are satisfied in the track generation process, reducing collisions with obstacles by 90\% compared to models without. 


\item Finally, we apply our diffusion models to generate track predictions for a single target, surpassing the performance of previous state-of-the-art (SOTA) models by an average of 9.2\% in terms of Average Displacement Error. We additionally set a new baseline on the challenging task of multi-target tracking in a large, partially observable domain.

\end{itemize}
\section{Related Works}
\subsection{Diffusion Models}
Deep diffusion models are a new class of generative models which model complex data distributions and have exploded in popularity within the computer vision community \cite{dhariwal2021diffusion, ho2022cascaded, croitoru2023diffusion}. As these models have shown promise on learning within high dimensional data manifolds, other research areas have begun to apply diffusion models as powerful generative and conditional generative models. In robotics, a key work by Janner et al. \cite{janner2022planning} shows that diffusion models can be used to generate plausible paths for planning. Diffusion policy \cite{chi2023diffusionpolicy} extends this to work to imitation learning by diffusing the action distributions to accomplish various pushing tasks. Recent contemporary work by Zhu et al. has also used cross-attention within diffusion models \cite{zhu2023madiff} to generate multi-agent tracks. However, their work assumes full observability which is not available in the adversarial tracking domain. We also take inspiration from image inpainting (reconstructing missing parts of an image) \cite{bertalmio2000image} to condition the diffusion sampling process on detections for producing better target track predictions. \emph{To the best of our knowledge, we are the first to utilize diffusion models for multi-target tracking under partial observability.}

\vspace{-1mm}
\subsection{Target Tracking}

Target tracking involves estimating the positions of one or more targets using sensor data \cite{souza2016target} and has various real-world applications such as surveillance \cite{gui2004power}, sports analysis \cite{gade2018constrained}, and traffic management \cite{kanistras2013survey}. Traditional approaches, like Particle Filters \cite{djuric2008target, rao2013visual, jia2016target} and Kalman Filters \cite{chen2000mixture, leven2009unscented}, dominate target tracking but require accurate knowledge or estimation of the target's dynamics model. However, recent advancements in model-free object tracking with images have emerged \cite{soleimanitaleb2019object}. Our work differs from prior work in computer vision as we rely on sparse observations or detected locations instead of images to predict future target trajectories.

\begin{figure*}[ht]
\centering

    \includegraphics[width=0.95\textwidth]{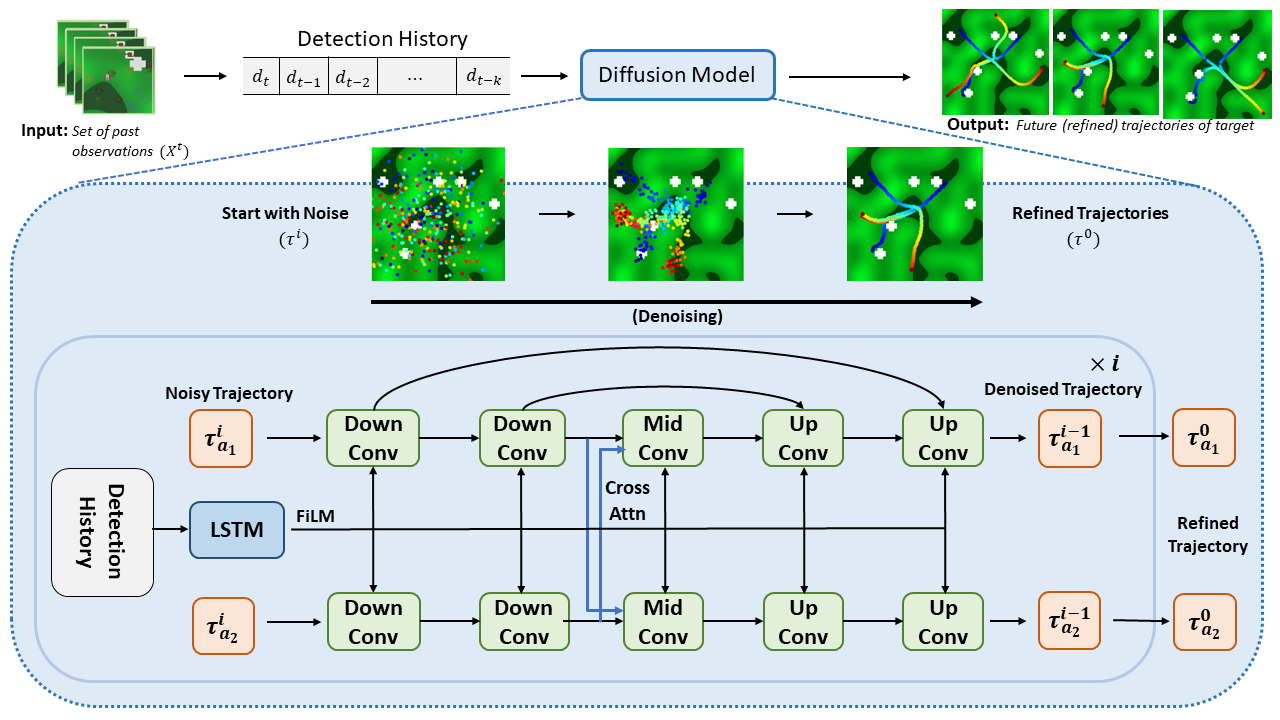}
    \caption{Our proposed architecture (\model) is a diffusion probabilistic model that iteratively refines trajectories for hypothesizing the future states of an adversarial target from a given history of past detections. Here we show cross attention across two parallel tracks but the model can be used for any number of $n$ agents. The color gradient shows the time horizon for each trajectory (blue to red).}
    \label{fig:architecture}
    \vspace{-2mm}
\end{figure*}

Adversarial tracking involves targeting an intelligent opponent trying to evade trackers \cite{nashed2022survey}. Previous works assume access to target states/observations for training predictive models \cite{he2016opponent, raileanu2018modeling, grover2018learning}. However, this assumption is unrealistic in large environments due to non-cooperative adversaries and a limited field of view. Prior work \cite{prev_work} introduced GrAMMI, which predicts dynamic target locations using partial observations from a team of trackers. However, it only focused on single target tracking and was unable to generate predictions for multiple time horizons. Our current work addresses these limitations by utilizing diffusion models to generate trajectories up to any time horizon and extending them to multi-target tracking.


\section{Background}
\subsection{Partially Observable Markov Game}
We define adversarial tracking as a Partially Observable Markov Game (POMG), which consists of a set of states $\mathcal{S}$, a set of private agent observations ${\mathcal{O}_1, \mathcal{O}_2, \ldots, \mathcal{O}_M}$, a set of actions ${\mathcal{A}_1, \mathcal{A}_2, \ldots, \mathcal{A}_M}$, and a transition function $\mathcal{T}: \mathcal{S} \times \mathcal{A}_1 \times \ldots \times \mathcal{A}_M \mapsto \mathcal{S}$ for M-agents. At each time step $t$, agents receive an observation $O_i^t\in\mathcal{O}_i$, choose an action $a_i^t \in \mathcal{A}_i$, and receive a reward $r_i^t$ based on the reward function $R: \mathcal{S} \times \mathcal{A}_i \mapsto \mathbb{R}$. The initial state is drawn from an initial state distribution $\rho$. 

We simplify the observation formulation of all agents to produce a single array of detections for all targets denoted as $\{d\}_{1...t}$. We denote the trajectory ($\tau$) of adversary states as $\tau_{n} \; \forall n \in N $, where $N$ is the number of targets being tracked. Thus, the goal of \smodel is to estimate the joint trajectory of all agents $p_\theta(\tau_{1...n} | \{d\}_{1...t})$.



In this work, we address multi-target and single-target tracking. For the multi-target tracking case, we ablate an assumption, \camera{where we either assume we know or \textit{do not} know the origin of a given detection. If we assume we know the detection origin, then we do not have to solve the data association problem. Otherwise, the model must perform target assignment to distinguish the paths.}



\subsection{Diffusion Probabilistic Models}
Diffusion models are a class of generative models that learn a target distribution through an iterative denoising process $p_\theta(x^{i-1} | x^i)$. The model learns how to reverse the forward noising process $q(x^i | x^{i-1})$, which is commonly parameterized as a Gaussian $\mathcal{N} \sim (0, I)$. Traditionally, $x$ is used to represent images but in our work, we replace this notation with $\tau$ as we are generating trajectories. 

The training process consists of a noising and denoising process. We utilize the Denoising Diffusion Probablistic Model (DDPM) formulation \cite{ho2020denoising} and create noisy trajectories with Equation \ref{eq:noise}, where $\bar{\alpha}^i$ is a noise scheduler dependent on the diffusion process timestep $i$. 
\begin{equation}
    \label{eq:noise}
    q(\tau^i | \tau^{0}) := \mathcal{N} (\tau^i ; \sqrt{\bar{\alpha}^i} \tau^{0}, {(1 - \bar{\alpha}^i}) \textbf{I})
\end{equation}
We learn a denoising network $\epsilon_\theta$ to predict the random noise at all denoising iterations $i$ (Equation \ref{eq:denoise}). 
\begin{equation}
    \label{eq:denoise}
    \mathcal{L} = MSE \left( \epsilon^i, \epsilon_\theta(\sqrt{\bar{\alpha}^i} \tau^0 + \sqrt{1 - \bar{\alpha}^i} \epsilon, i) \right)
\end{equation}
Finally, with a trained denoising network $\epsilon_\theta$, we can iteratively denoise a trajectory $\tau$ from pure Gaussian noise using Equation \ref{eq:sample}, which is equivalent to minimizing the negative log-likelihood of the samples generated by the model distribution under the expectation of the data distribution \cite{sohl2015}.
\begin{equation}
    \label{eq:sample}
    \tau^{i-1} = \frac{1}{\sqrt{\alpha^i}} \left( \tau^i - \frac{1 - \alpha^i}{\sqrt{1 - \bar{\alpha}^i}} \epsilon_{\theta}(\tau^i, i) \right)  + \mathcal{N} (0, \sigma^2 I) 
\end{equation}

\subsection{Domains and Target Behavior}
We test our models in the Prison Escape and Narco Traffic Interdiction (Smuggler) domains first described in \cite{prev_work}.

\subsubsection{Narco Traffic Interdiction}
This simulation involves illegal maritime drug trafficking along the Central American Pacific Coastline. A team of tracker agents, including airplanes and marine vessels, pursue a drug smuggler. Airplanes have a greater search radius and speed, while vessels can capture the smuggler. The smugglers must reach rendezvous points before heading to hideouts. The tracking team is unaware of hideout and rendezvous point locations. Episodes end when the smuggler reaches a hideout or is captured.

\subsubsection{Prisoner Escape}
In this scenario, a team consisting of cameras, search parties, and helicopters work together to track down an escaped prisoner, which is heading to one of several goal locations (hideouts).  The game takes place on a map measuring $2428 \times 2428$ units, with various mountains representing obstacles. The motivation behind this domain stems from situations encountered in military surveillance and border patrol, where the objective is to track and intercept adversaries.

Goal locations for the fugitive are randomly selected without replacement from a predetermined set for each episode. The Prison Escape scenario incorporates evasive behaviors for the prisoner and introduces a fog-of-war element that limits the detection range of all agents. Notably, the tracking agents are only capable of tracking the prisoner and not capturing them, allowing for analysis of long-term predictions. The episode concludes either when the prisoner reaches a goal location or after a maximum number of timesteps.



\section{Methodology}

Given sparse observations of the target $\{d\}_{1...t}$, our diffusion model generates possible paths of the various target(s) $\tau_n$. We describe \model's design architecture for the multi-target and single-target domain. 

\subsection{Model Architecture}
The multi-target diffusion model consists of two key components, 1) the temporal U-net and 2) the cross-attention mechanism between parallel tracks. 

\subsubsection{Temporal U-net}
We utilize a 1D temporal CNN-based architecture from \cite{chi2023diffusionpolicy} as our noise prediction network $\epsilon_\theta$ for each target agent in the environment. This architecture predicts an entire trajectory non-autoregressively and uses temporal convolutional blocks to encode the trajectory. Within the diffusion framework, the temporal U-net takes as input a noisy trajectory and outputs a refined (less noisy) trajectory. We repeat this $i$ times to produce a noiseless trajectory from the noisy one. For $a$ agents in the environment, we have use $a$ parallel denoising networks, where each pathway denoises the trajectory for a single agent.

We utilize Feature-wise Linear Modulation (FiLM) at each convolutional layer \cite{perez2018film} to condition the generative process on past detections $\{d\}_{1...t}$. The history of past detections $\{d\}_{1...t}$ is encoded in an LSTM, where each detection consists of the $\delta t, x, y$ denoting the time since detection and location of the detection. Figure \ref{fig:architecture} shows the full multi-agent denoising architecture.

\subsubsection{Cross Attention}
A key assumption in \smodel is that the track generation is permutation equivariant--- the ordering of the track inputs does not impact the results. This is achieved by sharing parameters between the track generators and using cross-attention to communicate information from one track to the other. The cross-attention formulation is a variant of the scaled dot product attention \cite{vaswani2017attention}:
\begin{equation}
    {x^m}' = \sum_m \text{softmax} \left( \frac{Q^m K^{n^T}}{\sqrt{dim_{k}} } \right) V^n
\end{equation}
where $Q \in \mathbb{R}^{N \times dim}, K \in \mathbb{R}^{N \times dim}, V \in \mathbb{R}^{N \times dim}$ are vectors of query, key, and value. $N$ is the number of query, key, and value vectors, $dim$ is the dimension of the vector, and $m, n$ are the indices for each target agent $A$. The attention module can be interpreted as a combination of both self-attention ($m=n$) and cross-attention across other agents ($m \neq n$). Crucially, the computation can be batched as the key, queries, and values $k, q, v$ for each agent track only needs to be computed once for each agent. Finally, we adopt the multi-headed attention formalism and concatenate multiple heads to produce ${x^m}'$. We use the cross attention embeddings by interspersing them between the convolutional blocks in the U-net. 

\subsubsection{Single-Target Architecture}
In the special case where we are only tracking a single agent, the cross-attention module is not used. In this formulation, a single history of detections is passed through the LSTM, and the model generates a single trajectory. 




    



\subsection{Constraint-Guided Sampling}
Within our domain, two dominant constraints exist: 1) the motion model of target agents and 2) obstacle (mountains) constraints. In this work, we adapt classifier-based guidance \cite{dhariwal2021diffusion}, which was first used in image diffusion models to steer models towards certain classes. \edit{Classifier-based guidance uses a trained discriminative model to estimate $p(y|x)$, which denotes a class of an image based upon its input.} The guidance augments the diffusion sampling procedure by changing the predicted means using the gradients of the classifier $\nabla_x \text{log} p(y|x)$. 

We adapt \textit{classifier}-based sampling to \textit{constraint}-based sampling by substituting the classifier with an objective function $ J(\mu^i)$. We denote the mean of the trajectory we learn as $\mu$ and the sampled path as $\tau$. Then, the guidance process can be written as $\tau^{i-1} \sim \mathcal{N}(\tau^{i}; \mu_\theta(\tau^i) + s \Sigma g, \Sigma)$, where the mean of the new distribution is perturbed by $g = \nabla_\tau J(\mu)$ and $s$ is a gradient scale. In our implementation, we use an additional Adam optimizer to perform this gradient update (\textbf{Algorithm \ref{algo:sampling}}). Using the \edit{Adam optimizer \cite{KingmaB14}} relieves the need for hyper-parameter tuning $s$ and allows us to combine multiple constraints together.

\begin{algorithm}[h]
\caption{Optimizer Based Constraint-Guided Sampling}
\begin{algorithmic}[1]
\STATE \textbf{Input:} constraint function $J(\mu_\theta)$ 
\STATE $\tau^T \leftarrow$ sample from $\mathcal{N}(0, I)$ \label{line:noise}
\FOR{all $i$ from $T$ to $1$} \label{line:for}
    \STATE $(\mu^i, \Sigma^i) \leftarrow \mu_{\theta}(\tau^{i}), \Sigma_{\theta}(\tau^{i})$ \label{line:model}
    \FOR{each gradient step} \label{line:step}
        \STATE $\mu^i \leftarrow \mu^i + \lambda \nabla_{\mu^i} J(\mu^i)$ \label{line:constrain}
    \ENDFOR
    \STATE $\tau^{i-1} \sim \mathcal{N}(\mu^i, \Sigma^i)$ \label{line:sample}
    \STATE $\tau_0 \leftarrow s_0$ if $s_0$ is known \label{line:condition}
\ENDFOR
\STATE \textbf{Return} $\tau^0$
\end{algorithmic}
\label{algo:sampling}
\end{algorithm}

In our modified sampling algorithm (\textbf{Algorithm \ref{algo:sampling}}), we begin with a completely noisy trajectory (line \ref{line:noise}). Then, the denoising process occurs for $T$ timesteps (line \ref{line:for}), where the denoised trajectory means are sampled from the model (line \ref{line:model}). We then use our constraint functions to move the means (line \ref{line:constrain}) and sample from the new distribution (line \ref{line:sample}). Finally, we condition the model on detected locations at each diffusion timestep (line \ref{line:condition}).  

We use two constraint functions, one for the motion model and the second for the obstacles. 

\begin{enumerate}
    \item \textbf{Motion Model Constraint}: $\sum_t{\lVert \tau_t - \tau_{t+1}  \rVert}$ \\ 
    For each consecutive point in our trajectory, we create a simple smoothness loss such that consecutive states in the trajectory are close by.
    \item \textbf{Obstacle Constraint}: $\lVert \tau_t - c \rVert < \epsilon \hspace{1mm} \forall t \in T, c \in C$ \\
    For each state $(\tau_t)$ in the trajectory and each obstacle $C$ on the map, we provide a loss that pushes states away from obstacles in the environment.
\end{enumerate}

\subsection{Conditioning Detected Observations}

While almost all the detected state information about the adversary is in the past, we provide a way to implement detected locations at the current time horizon $t=0$ directly into the diffusion model sampling process. We alter the sampling process of the diffusion model, where if the detected location at the current timestep is known, we replace the sampled value with the known location after each diffusion timestep $i$ (Algo \ref{algo:sampling}, line \ref{line:condition}). 

Planning with diffusion models have used a similar process to goal-condition trajectories based on the starting and ending location \cite{janner2022planning}. This solution is inspired by inpainting \cite{sohl2015, lugmayr2022repaint} in computer vision, where parts of an image are known and the diffusion model must generate the rest of the image. 

\section{Evaluation}
We evaluate our single-target models for target tracking in the Prison Escape and Smuggler scenario introduced in \cite{prev_work} and use Monte-Carlo Sampling from the trained diffusion model to estimate the distribution of trajectories by generating 30 paths per sample. 
We use three datasets in the Prison Escape scenario (Prisoner-Low, Prisoner-Medium, Prisoner-High) that contain opponent detection rates of $12.9\%$, $44.0\%$, and $63.1\%$, respectively and two Smuggler datasets with opponent detection rates, $13.8\%$ and $31.5\%$. 

We create new multi-agent datasets within the same domain. However, we do not include target adversaries in this domain for simplicity. Instead, we randomly sample 10-12\% of the timesteps and assume that these are the detected locations. These detected location samples are not resampled during training such that there is only one set of detections per trajectory rollout. We create two types of behavior 1) where the all the agents meet together before traveling to the goal location and 2) where the agents directly go to the goal location. This assumption produces a multimodal distribution where the behaviors of the agents in the domain are dependent on each other. All agents use $A^*$ to traverse through the landscape, where terrain with lower coverage visibility is preferred over higher coverage visibility. In the maps shown in Figure~\ref{fig:multiagent_paths}, the low visibility areas correspond to the darker regions of the map. In our environment, we choose three target agents to track but our model can be used to track any $n$ number of targets.


In our analysis, we examine two scenarios regarding detections. The first scenario assumes that we have knowledge about the origin of each detection. The second scenario assumes that we \textit{do not} have any information about the origin of the detections. In real-world target tracking situations, it is common for us to lack knowledge about the origin of the detections and, therefore, implicitly conduct target assignments.

\subsection{Metrics}
We evaluate \smodel on two measures used in prior work \cite{zhu2023madiff} --- Average Displacement Error (ADE, minADE) to compare against previous models. In the case of multi-agent tracking, we average the metrics across all agents. Previous work that fit a probability distribution included log-likelihood $\text{log}(p(s_t|\theta))$, as a measure, Where $\theta$ is the model parameters. Computing the exact log-likelihood through the diffusion process is still an open research topic \cite{song2021scorebased}, where deterministic samplers based on probablistic flow ODEs have been used to compute exact likelihoods.


\begin{enumerate}
    \item \textbf{Average Displacement Error (ADE)}: Given a ground truth trajectory $\tau$, we compute the average $l_2$ distance between each sampled trajectory and the ground truth trajectory over all timesteps.

    \item \textbf{Minimum Average Displacement Error (minADE)}: minADE measures the distance of the closest sampled path to the ground truth trajectory.
\end{enumerate}

\section{Results \& Discussion}

We evaluate our multi-target models within the multi-target Prison Escape domain, showing ablations for knowing the origin of the detections and without knowing the origins of the detections.

The performance of our single-target tracking model was evaluated on the three Prison Escape datasets introduced in \cite{prev_work}. Our models show better ADE than the previous best Gaussian Mixture-based model (GMM) on every prediction horizon.

\subsection{Multi-Target Tracking}
We present findings regarding the performance of models with known detection origins compared to those without, as well as the qualitative analysis of the generated tracks in the multi-target tracking domain.



\begin{figure}[htbp]\centering
  \includegraphics[width=0.99\columnwidth, keepaspectratio]{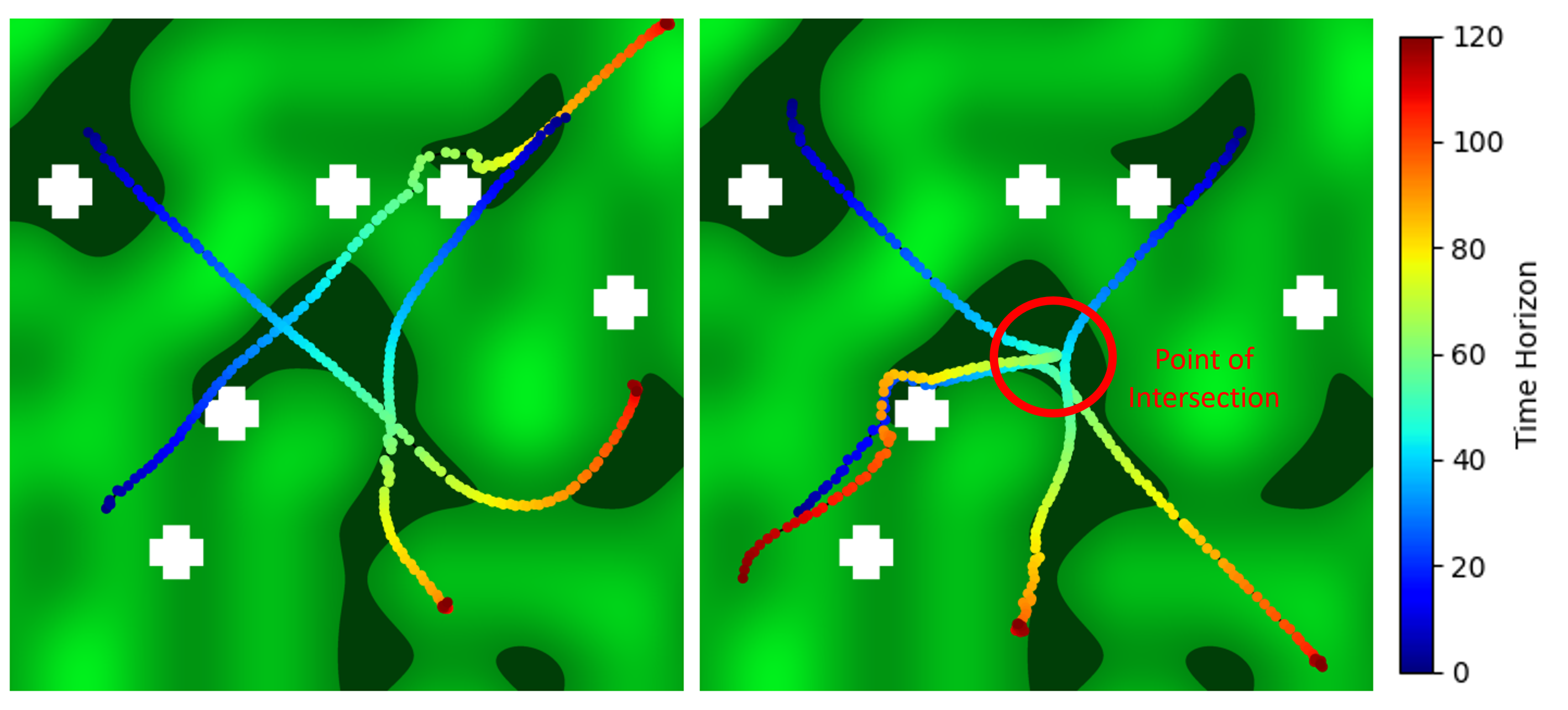}
  \caption{Multi-Hypothesis Trajectory Samples: In one sample (left) the agents take separate paths to the final location while the other sample (right) shows all agents meeting at an intermediary point. }
  \label{fig:multiagent_paths}
  \vspace{-1mm}
\end{figure}

First, we are able to qualitatively show the multimodal behavior and dependencies between tracks with the cross-attention mechanism (Figure \ref{fig:multiagent_paths}). Given the same inputs, the model produces tracks that do not intersect (left image) along with tracks that do intersect (right image), showing the models have learned the relationship between the tracks that is inherent within the dataset. 

\begin{figure}[h]
    \centering
    \includegraphics[width=0.95\columnwidth, keepaspectratio]{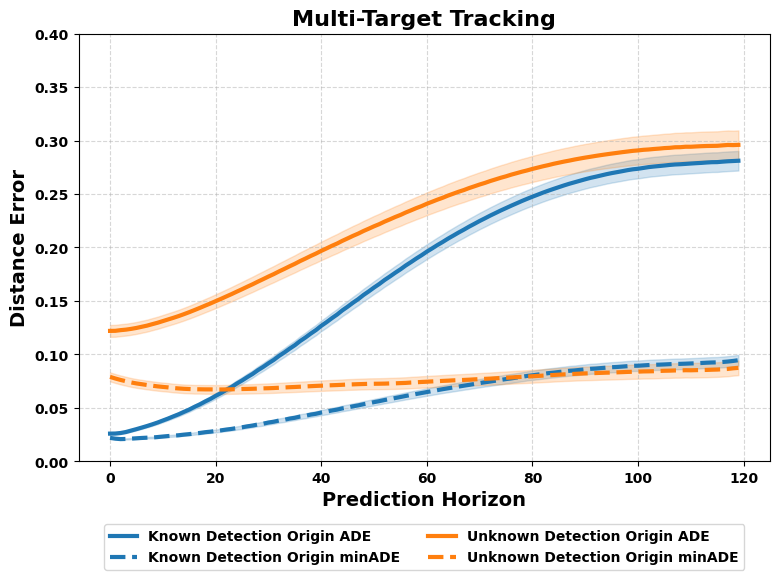}
    
    \caption{Detection Origin Ablation: ADE and minADE over Prediction Horizon. Shaded regions represent standard error of the mean.}
    \label{fig:ade_multi}
\end{figure}

\textbf{Detection Origin Ablation}: 
We compare the ADE and minADE for models with detection origin information and those without (Figure \ref{fig:ade_multi}). For models with origin information, we pass each target's detections through its own LSTM encoder to retrieve an embedding per target. This embedding is fed uniquely into each of the track generators. For models without detection origin information, we feed all detections through a single LSTM encoder whose embedding is shared amongst the diffusion tracks. 

As expected, the models that possess knowledge of detection origins exhibit superior performance over all time horizons in terms of ADE and minADE compared to the models lacking such knowledge. Notably, there is a significant difference between ADE and minADE, indicating that while the distribution of paths is large, we can also generate hypothetical tracks that align well with the actual trajectories.

Furthermore, we observed that models without detection origins exhibit significantly higher error rates comparatively for shorter prediction horizons compared to longer horizons. As the prediction horizon increases, trajectories are drawn to the discrete set of total goal locations, and the diffusion model directs trajectories towards these points. Therefore, in our domain, the challenge of assigning targets to the detections becomes more evident in shorter prediction horizons. 

\smodel is the first method capable of performing multi-target tracking under partial observability and implicitly conducts target assignment to generate consistent interactions between agents.



\subsection{Constraint-Guided Samples}

\begin{figure}[htbp]
  \centering
  \includegraphics[width=0.98\columnwidth, keepaspectratio]{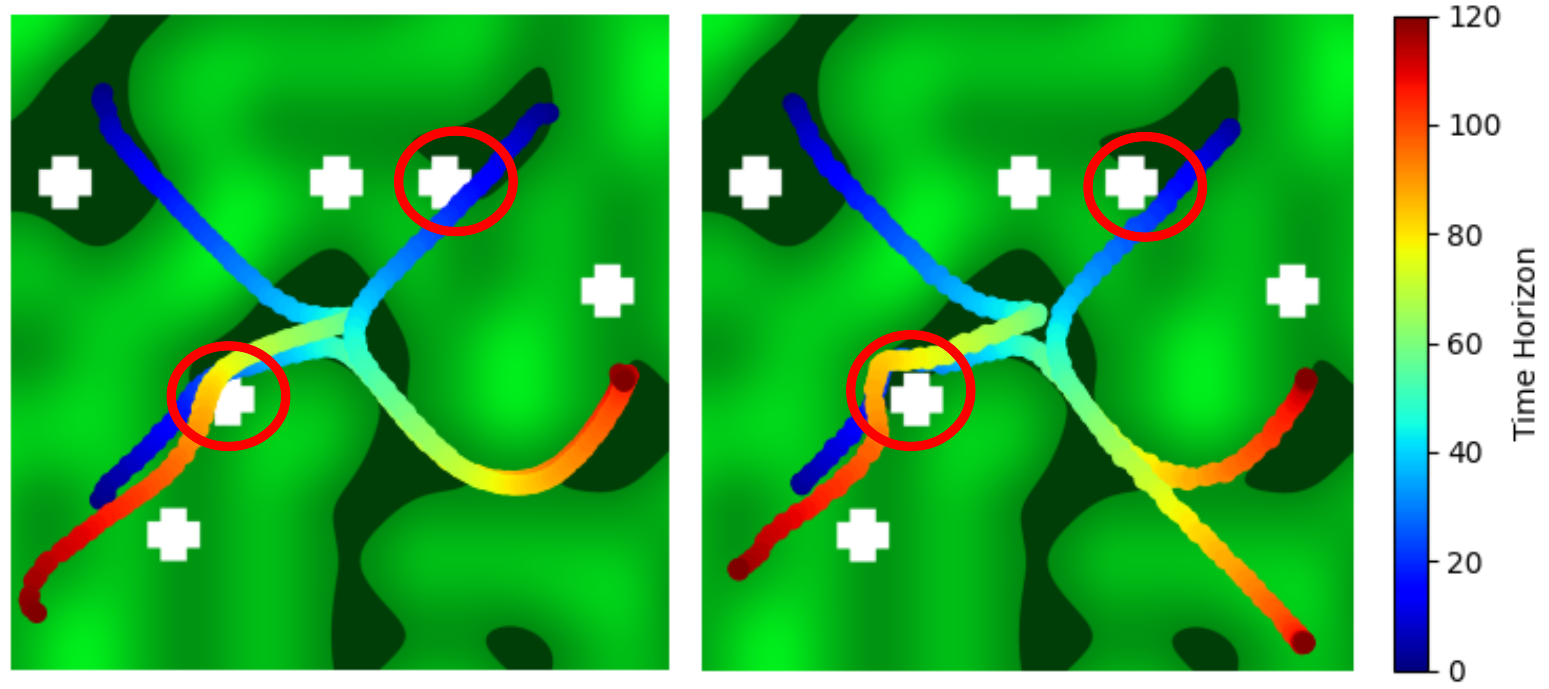}
    
  \caption{Original sampling (left) vs Constraint-Based sampling (right): We show key improvements to avoid the mountains in the sample on the right (circled in red).}
  \label{fig:mountain_comparison}
  \vspace{-5mm}
\end{figure}

We compare how effective the constraint-guided sampling process prevents collisions with obstacles in the environment. A visualization of the difference is shown in Figure \ref{fig:mountain_comparison}. We report our findings where samples generated without the constraint-sampling method resulted in an average of 5\% of the produced states colliding with obstacles. On the other hand, states generated using the constraint-sampling method only encountered mountain collisions approximately 0.641\% of the time. Our findings demonstrate that the new sampling procedure led to a significant 90\% decrease in the number of collisions in the generated trajectories and produces more consistent hypothetical trajectories with actual trajectories.

\subsection{Single-Target Tracking}
We present our results on single-target tracking using a unique version of the model without cross-attention, as there is only one agent. Our findings for single target tracking are presented based on the analysis of three Prisoner Escape and Smuggler datasets (Table \ref{tab:prisoner}). The results demonstrate that diffusion models outperform the previous best Gaussian Mixture models on average by 9.2\%. At higher prediction horizons of 60, 90 and 120 timesteps into the future, the diffusion models yield improvements of 12.2\%. Additionally, our models are able to generate complete trajectories compared to the previous models which could only predict states at a fixed horizon length.

The non-parametric formulation of the diffusion model lends itself better for trajectory generation in our sparse detection environments than fitting a mixture of Gaussians. Fitting a mixture of Gaussians requires identifying the appropriate number of Gaussians --- a hyperparameter for the target tracks. The diffusion models overcome this requirement and are able to represent a more diverse set of tracks. 

By incorporating the complete trajectories generated by our diffusion models, we can enhance the capabilities of searching agents and improve target containment strategies. The availability of full track predictions allows us to anticipate the target's movements, identify potential escape routes, and strategically position agents to cut off those routes effectively. This enables us to employ more advanced policies for cornering the target and maximizing the chances of capture.



\setlength{\tabcolsep}{3pt}
\newcommand{\ra}[1]{\renewcommand{\arraystretch}{5}}

\begin{table}[] \centering
\begin{tabular}{c|l|rrrrr}
\multicolumn{2}{l}{} & \multicolumn{5}{c}{\textbf{Prediction Horizon}} \\
\multicolumn{1}{l}{}             &                  & 0 min           & 30 min         & 60 min         & 90 min         & 120 min        \\
\toprule
\multirow{3}{*}{\rotatebox[origin=c]{90}{P-Low}} 
& VRNN             & {0.106}  & 
                                 {0.093} & 
                                 {0.119} & 
                                 {0.146} & 
                                 {0.177} \\
& \oldmodel w/o MI     & 0.060            & 0.083          & 0.109          & 0.144          & 0.165          \\
                                 & \oldmodel & 0.060            & 0.080           & 0.110           & 0.154          & 0.163          \\
                                 & Ours             & \textbf{0.057}  & \textbf{0.077} & \textbf{0.100}   & \textbf{0.127} & \textbf{0.154} \\

\bottomrule
\multirow{3}{*}{\rotatebox[origin=c]{90}{P-Med}} & 
VRNN             & {0.172}  & 
                                 {0.086} & 
                                 {0.110} & 
                                 {0.144} & 
                                 {0.167} \\ &
\oldmodel w/o MI     & 0.047           & 0.078          & 0.110          & 0.142          & 0.168          \\
                                 & \oldmodel & 0.049           & 0.077          & 0.110           & 0.146          & 0.167          \\
                                 & Ours             & \textbf{0.046} & \textbf{0.076} & \textbf{0.103} & \textbf{0.129} & \textbf{0.153} \\

\bottomrule
\multirow{3}{*}{\rotatebox[origin=c]{90}{P-High}}  
& VRNN             & {0.105}  & 
                                 {0.059} & 
                                 {0.100} & 
                                 {0.117} & 
                                 {0.145} \\
& \oldmodel w/o MI     & 0.016           & 0.057          & 0.095          & 0.132          & 0.167          \\
                                 & \oldmodel & \textbf{0.015}          & 0.056          & 0.092          & 0.122          & 0.162          \\
                                 & Ours             & 0.017  & \textbf{0.054} & \textbf{0.078} & \textbf{0.099} & \textbf{0.118} \\
\bottomrule
\toprule

\multirow{3}{*}{\rotatebox[origin=c]{90}{S-Low}}    
& VRNN             & {0.147}  & 
                                 {0.156} & 
                                 {0.186} & 
                                 {0.187} & 
                                 {0.203} \\
& \oldmodel  w/o MI    & 0.122           & 0.142          & 0.159          &   0.169  &       0.182    \\
                                 & \oldmodel & 0.121           & 0.144          & 0.181          &   0.183       &    0.193      \\
                                 & Ours & \textbf{0.112}  & \textbf{0.123} & \textbf{0.135} & \textbf{0.148}  & \textbf{0.160}      \\
\bottomrule
\multirow{3}{*}{\rotatebox[origin=c]{90}{S-High}}   
& VRNN             & {0.138}  & 
                                 {0.153} & 
                                 {0.183} & 
                                 {0.179} & 
                                 {0.185} \\
& \oldmodel  w/o MI    & 0.125           & 0.144          & 0.161          &   0.169      &    0.178    \\
                                 & \oldmodel & 0.131           & 0.163          & 0.174      &  0.175    &   0.184          \\
                                 & Ours & \textbf{0.113}  & \textbf{0.124} & \textbf{0.138} & \textbf{0.152}  & \textbf{0.163}      \\
\bottomrule
\end{tabular}

\caption{ADE Results for three Prisoner Escape (P-low, P-med, P-high) and two Smuggler (S-low, S-high) Datasets. Bolded values represent the best-performing model.}
\label{tab:prisoner}
\vspace{-3mm}
\end{table}

\section{Limitations}
While our diffusion models show great improvements over previous models, a main limitation is the sampling time for generating tracks. Improving sampling speeds for diffusion models is currently an active area of research \cite{salimans2022progressive}.

Additionally, our model can perform track prediction for future timesteps but implicitly learns the target track assignment from past detections. We, therefore, did not consider the task of track \textit{reconstruction} in this work, where all trajectories generated are of future timesteps and not of the past. Finaly, we currently assume all agents are homogeneous in this work but could extend this to heterogeneous agents by removing the shared weights between track generators. 


\section{Conclusion}

We proposed a novel approach using diffusion probabilistic models for single and multi-target tracking in large-scale environments. Our model incorporates cross-attention, constraint-guided sampling, and conditioning techniques to improve track prediction accuracy and adhere to motion model and environmental constraints. The experimental results demonstrate the effectiveness of the approach, surpassing the performance of previous state-of-the-art models in single-target tracking and achieving successful multi-target tracking with improved target track assignment.

\bibliographystyle{plain}
\bibliography{ref.bib}

\end{document}